\documentclass[12pt,a4paper]{article}
\usepackage[latin1]{inputenc}
\usepackage[T1]{fontenc}
\usepackage{pgf}
\usepackage{url}
\usepackage[english]{babel}
\usepackage{multirow}
\usepackage{fancyhdr}
\usepackage{anysize}
\usepackage{amssymb}
\usepackage{graphicx}
\usepackage[latin1]{inputenc}
\usepackage{url}
\usepackage{algorithm}
\usepackage{algpseudocode}
\usepackage[sort, numbers]{natbib}
\usepackage{ntheorem}
\usepackage{authblk}
\begin{document}


\makeatletter
\let\@fnsymbol\@arabic
\makeatother

\title{\bf{Event and Anomaly Detection Using Tucker3 Decomposition}}
\author{Hadi Fanaee-T \thanks{LIAAD-INESC TEC, FEUP- University of Porto, hadi.fanaee@fe.up.pt} , Márcia Oliveira \thanks{LIAAD-INESC TEC, FEP- University of Porto, marcia@liaad.up.pt} , João Gama \thanks{LIAAD-INESC TEC, FEP- University of Porto, jgama@fep.up.pt}  \\ Simon Malinowski \thanks{INESC TEC, FEUP- University of Porto, sjfm@inescporto.pt} , Ricardo Morla \thanks{INESC TEC, FEUP- University of Porto, ricardo.morla@inescporto.pt}}

\maketitle

\begin{abstract}

Failure detection in telecommunication networks is a vital task. So far, several supervised and unsupervised solutions have been provided for discovering failures in such networks. Among them unsupervised approaches has attracted more attention since no label data is required. Often, network devices are not able to provide information about the type of failure. In such cases the type of failure is not known in advance and the unsupervised setting is more appropriate for diagnosis. Among unsupervised approaches, Principal Component Analysis (PCA) is a well-known solution which has been widely used in the anomaly detection literature and can be applied to matrix data (e.g. Users-Features). However, one of the important properties of network data is their temporal sequential nature. So considering the interaction of dimensions over a third dimension, such as time, may provide us better insights into the nature of network failures. In this paper we demonstrate the power of three-way analysis to detect events and anomalies in time-evolving network data.
\end{abstract}


{\bf Citation :} Hadi Fanaee-T et al., Event and Anomaly detection Using Tucker3 Decomposition, In Proceedings of 20th European Conference on Artificial Intelligence (ECAI'2013)- Ubiquitous Data Mining Workshop, pp. 8-12, vol. 1 , August 27-31, 2012, Montpellier, France.

\section{Introduction}\label{sec:introduction}
Event detection can be briefly described as the task of discovering unusual behavior of a system during a specific period of the time. On the other hand, anomaly detection concentrates on the detection of abnormal points. So clearly it is different from event detection since it just considers the points rather than a group of points. Our work takes into account both issues using multi-way data analysis. Our methodology comprises the following steps: 1) Anomaly detection: detection of individual abnormal users 2) Generating user trajectories (i.e. behavior of users over time), 3) Clustering users' trajectories to discover abnormal trajectories and 4) Detection of events: group of users who show abnormal behavior during specific time periods. Although there is a rich body of research on the two mentioned issues, to the best of our knowledge we are the first ones applying multi-way analysis to the anomaly and event detection problem. In the remainder of this section we explain some basic and related concepts and works. Afterward, we define the problem, and then discuss three-way analysis methods. Hereafter, we introduce the dataset and experiments. Finally, we discuss the results and point out possible future directions.

\subsection{Anomaly Detection} \label{sec:anomaly}

Anomaly is as a pattern in the data that does not conform to the expected behavior \cite{r1}. Anomaly detection has a wide range of application in computer network intrusion detection, medical informatics, and credit card fraud detection. A significant amount of research has been devoted to solve this problem. However our focus is on unsupervised methods. Anomaly detection techniques can be classified into five groups \cite{r1}: classification-based, clustering-based, nearest neighbor based, statistical methods, information theory-based methods and spectral methods. Based on this classification, our method is placed in the group of spectral methods. These approaches first decompose the high-dimensional data into a lower dimension space and then assume that normal and abnormal data points appear significantly different from together. This some benefits: 1) they can be employed in both unsupervised and supervised settings 2) they can detect anomalies in high dimensional data, and 3) unlike clustering techniques, they do not require complicated manual parameter estimation. So far, most of the work related to spectral anomaly detection was based on Principal Component Analysis (PCA) and Singular Value Decomposition (SVD). Two of the most important applications of PCA during recent years has been in the domain of intrusion detection \cite{r2,r3} and traffic anomaly detection \cite{r4,r5}.

\subsection{Event Detection} \label{sec:anomaly}

Due to huge amount of sequential data being generated by sensors, event detection has become an emerging issue with several real-world applications. Event is a significant occurrence or pattern that is unusual comparing to the normal patterns of the behavior of a system \cite{r6}. This can be natural phenomena or manual system interaction. Some examples of events can be an attack on the network, bioterrorist activities, epidemic disease, damage in an aircraft, pipe-breaks, forest fires, etc. A real system behaves normally most of the time, until an anomaly occurs that may cause damages to the system. Since the effects of an event in the system are not known a priori, detecting and characterizing abnormal events is challenging. This is the reason why most of the time we cannot evaluate different algorithms. One solution might be injection of artificial event into the normal data. However, construction of a realistic event pattern is not trivial \cite{r7}. 
 
\subsection{Hidden Markov Models} \label{sec:hmm}

Hidden Markov Models (HMMs) have been used at least for the last three decades in signal processing, especially in domain of speech recognition. They have also been applied in many other domains as bioinformatics (e.g. biological sequence analysis), environmental studies (e.g. earthquake and wind detection), and finance (financial time series). HMMs became popular for its simplicity and general mathematical tractability \cite{r8}.

HMMs are widely used to describe complex probability distributions in time series and are well adapted to model time dependencies in such series. HMMs assume that observations distribution does not follow a normal distribution and are generated by different processes. Each process is dependent on the state of an underlying and unobserved Markov process \cite{r7}. Markov process denotes this property that given the value $X_t$ the values of $X_n$, $n \succ t$, do not depend on the values $X_m$, $m \prec t$. Using notations of \cite{r9} let:

\begin{itemize}
	\item T = Length of the Observation sequence
	\item N = Number of states
	\item Q = {$q_0$, $q_1$, ... , $q_{(N-1)}$ } = distinct states of Markov process
	\item A = State transition probabilities
	\item B = set of N observation probability distributions
	\item $\pi$ = Initial State Distribution
	\item O = ($O_0$, $O_1$, ... , $O_{(T-1)}$) = observation sequence
\end{itemize}

A HMM Model is defined with the triple of $\lambda= (A, B, \pi)$.  It assumes that Observations are drawn using the observation probability distribution associated to the current state. The transition probabilities between states are given in matrix A.

The three main problems related with HMMs are the following. The first problem consists in computing the probability P(O) that a given observation sequence O is generated by a given HMM $\lambda$. The second problem consists in finding the most probable sequence of hidden states given an observation sequence O and $\lambda$  and the third problem is related to parameter inference. It consists in estimating the parameters of the HMM $\lambda$ that best fits a given observation sequence O. The mainly used algorithms to solve these problems are given in the last column of Table \ref{tab:tab1}. More details about these algorithms can be found in \cite{r10}. In this paper, we deal with the third problem to estimate the HMM parameters that best describe time series, as it will be explained in Section 2.

\begin{table}[ht]
\begin{center}
\caption{Three HMM Problems}\label{tab:tab1}
		\begin{tabular}{| l | l | l | l |}
    \hline
    \textbf{Problem} & \textbf{Input}  & \textbf{Output} & \textbf{Solution} \\ \hline
    Problem 1 & $\lambda, O$  & P(O) & Forward Backward algorithm\\ \hline
    Problem 2 & $\lambda, O$  & Best Q & Viterbi algorithm \\ \hline
    Problem 3 & O &  $\lambda$   & Baum-Welch algorithm\\ \hline
    \end{tabular}
\end{center}
\end{table}

\subsection{Three-way data analysis} \label{sec:threeway}

Traditional data analysis techniques such as PCA, clustering, regression, etc. are only able to model two dimensional data and they do not consider the interaction between more than two dimensions. However, in several real-world phenomena, there is a mutual relationship between more than two dimensions (e.g. a 3D tensor ($Users \times Features \times Time$)) and thus, they should be analyzed through a three-way perspective. Three-way analysis considers all mutual dependencies between the different dimensions and provides a compact representation of the original tensor in lower-dimensional spaces. The most common three-way analysis models are Tucker2, Tucker3, and PARAFAC \cite{r10} which are generalized versions of two-mode principal component model or, more specifically, SVD. Following, we briefly introduce Tucker3 model as the best-known method for analysis of three-way data.

\subsubsection{Tucker3 Model} \label{sec:tucker}

\begin{figure}[ht]
 \begin{center}
	\includegraphics[width=0.5\textwidth]{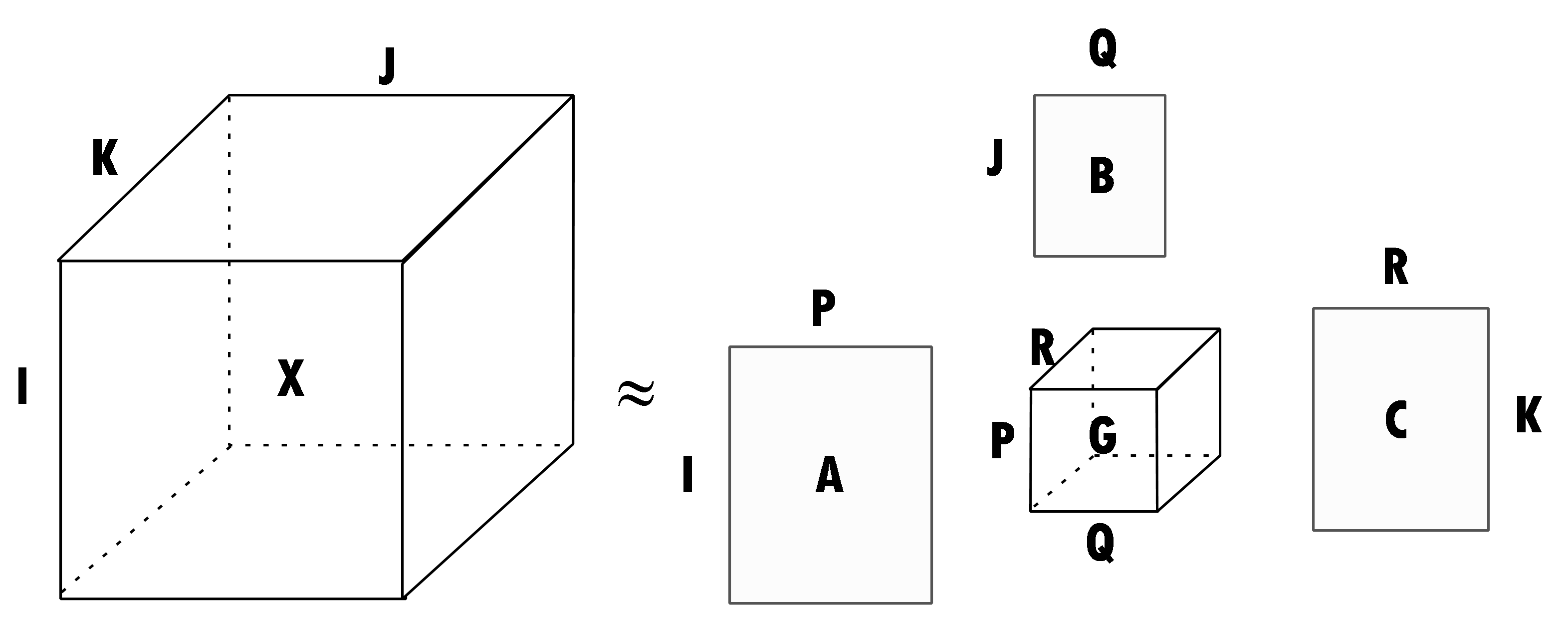}
 \end{center}
 \caption{Tucker3 Decomposition} \label{fig:f1}
\end{figure}

The Tucker3 model decomposes a three-mode tensor X into set of component matrices A, B, C and a small core tensor G. The following mathematical equation reveals the decomposition:

\begin{equation} \label{eq:tucker3}
X_{ijk} \simeq \sum_{p=1}^{P}\sum_{q=1}^{Q}\sum_{r=1}^{R}G_{pqr}\times a_{ip}\times b_{jq}\times c_{kr}
\end{equation}

Where P, Q and R are parameters of the Tucker3 model and represent the number of components retained in the first, the second and the third mode of the tensor, respectively.  This decomposition is illustrated in Figure \ref{fig:f1}.

\section{Problem definition} \label{sec:problem}

One of significant issues in telecommunication systems, such as IP/TV, is to detect the anomalies at both network and user level. In order to study this, target users are usually equipped with a facility in their modem which sends an automatic notification message to the central server when the connection of a client in the network is lost or reestablished. These modems are ubiquitous and geographically dispersed.

The modeling of such behavior is not straightforward because the number of notification messages is not equal for each user during the time period under analysis. For instance, one user may face 40 connection problems in an hour, hence generating 40 messages, while others may face 5 or even no problems at all. 

\begin{figure}[ht]
 \begin{center}
	\includegraphics[width=0.5\textwidth]{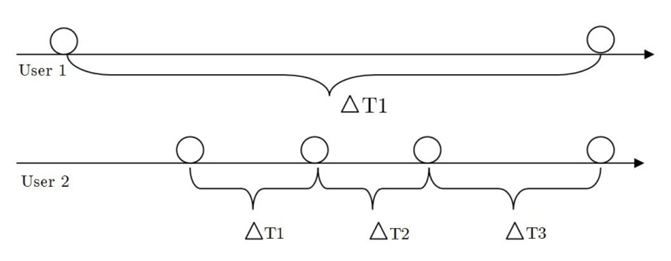}
 \end{center}
 \caption{Two sample users with different number of messages and different intervals} \label{fig:f2}
\end{figure}

In standard event detection problems, for each time point there is a measurement via one or multiple sensors. In the context of our application, such measurements do not take place at regular time points, since user modems (or sensors) only send messages to the server when something unexpected occurs. Figure \ref{fig:f2} illustrates two sample users. Each circle represents the time stamp at which a notification relative to the given user is received, while $\Delta T$ represents the inter-arrival time between two consecutive messages. As it can be seen, 2 messages were related to user 1 in that period, while 4 were related to user 2 during the same period. Also, the $\Delta T$ between messages is larger for user 1 than for user 2. This means that user 2 sent messages more frequently than user 1.  As in many other event detection problems, we could easily use the number of events per hour (measurement) at different users (sensors) to detect the events but this way we would lose the information content provided by the $\Delta T$'s. 

As the number of $\Delta T$  is not the same for each user, this feature cannot be directly integrated in our model. Hence, this would cause some vectors to have different lengths, which is not supported by the Tucker3 analysis. To solve this, every time-series of $\Delta T$  relative to a given user is modeled by a 2-state HMM obtained by the Baum-Welch algorithms \cite{r11}. 6 parameters are extracted from the HMM and are used to describe the time-series of $\Delta T$  of the users. Using this approach we obtain the same number of features for each user and, then, include this information in our feature vectors.

\begin{table}[ht]
\begin{center}
\caption{Datasets in tensor format}\label{tab:tab2}
		\begin{tabular}{| l | l | l | l |}
    \hline
    Data  & 1st mode (I): Users  & 2nd mode (J): Features & 3rd mode (K): Hours \\ \hline
    X102 & 102  & 10 & 720\\ \hline
    X909 & 909  & 10 & 720 \\ \hline
    \end{tabular}
\end{center}
\end{table}

\section{Data set} \label{sec:dataset}

Dataset is extracted from the usage log of a European IP/TV service provider. The raw dataset includes the notification messages of users in each line including their occurrence time. As previously mentioned, it is not possible to use this data directly in our modeling approach, so some pre-processing steps were performed. In addition to the obtained HMM parameters for each hour and for each user, we included another features, such as mean, variance, entropy and number of messages per hour, to our feature vector. We generated two separated datasets, each one spanning a time period of one month, which is equivalent to 720 hours. In one set we selected 102 users and in another we selected 909 users. The latter dataset is an extended version of the former. We then transformed both datasets to the tensor format.  These datasets are shown in a format of Tucker3 input tensor (Figure \ref{fig:f1}) in Table \ref{tab:tab2} where I, J, K represent users, features and hours modes, respectively.

\section{Experiments} \label{sec:experiments}

This section is divided into three subsections, according to the steps mentioned in the Introduction section. In subsection 1, we explain how we detect the abnormal users. In the next subsection we describe how we generate user trajectories And in the last subsection we explain how we cluster the trajectories using hierarchical clustering and detect events using user trajectories. 

\subsection{Abnormal Users} \label{sec:abnormalusers}

We applied Tucker3 model to both datasets X102 and X909 by employing a MATLAB package called Three-mode component analysis (Tucker3) \cite{r10}. Before that, we performed ANOVA test \cite{r10} to see the significance of three-way and two-way interaction in the data. The results of this test are presented in Table \ref{tab:tab3}. ANOVA Max 2D represents the maximum value obtained via different combinations of two-way modeling (e.g. I-J, J-K, I-K). As it can be seen, bigger numbers are obtained for three-dimension interaction (ANOVA 3D), which reveals that there is a mutual interaction between the three dimensions in both datasets that can be explained better with three-way modeling like Tucker3, than with two-way modeling like PCA.  

\begin{table}[ht]
\begin{center}
\caption{ANOVA test and selected model parameters P-Q-R}\label{tab:tab3}
		\begin{tabular}{| l | l | l | l | l |}
    \hline
    Data  & Anova Max 2D & ANOVA 3D & Selected Model P-Q-R & fit \\ \hline
    X102 & 26.18\%  & 38.90\% & 3-2-2 & 42.00\\ \hline
    X909 & 17.02\%  & 78.04\% & 40-2-4 & 51.01 \\ \hline
    \end{tabular}
\end{center}
\end{table}

The next step is to estimate the best parameters P, Q, R of Equation \ref{eq:tucker3}. P-Q-R is similar to what we have in PCA. In PCA we just determine the number of PCs for one dimension but here we need to determine the number of principal components for each one of the three modes. P, Q and R can assume values that fall within the interval [1,max], where max denotes the maximum number of entities in the corresponding mode. For example, in terms of  X102 the P-Q-R can go from 1-1-1 to 102-10-720.  These parameters are chosen based on a trade-off between model parsimony, or complexity, and goodness of fit. For instance, regarding the mentioned dataset, 1-1-1 gives about 28\% fit (less complete and less complex) and model 102-10-720 gives 100\% fit (most complete and most complex). If we try parameters 3-2-2 the model has a 42\% fit. So it can be more reasonable choice because it finds a good compromise between complexity and fit. In \cite{r10} the scree test method is proposed as a guideline to choose these parameters. We used this test to determine the best model for both datasets. The selected model parameters and their corresponding fits are presented in Table \ref{tab:tab3}. This means that, for example, for dataset X102 if we choose Tucker3 model with 3, 2 and 2 components to summarize the users, features and hours modes, respectively, the model is able to explain 42\% of the total variance contained in raw data.

After the estimation of model parameters, we used the selected model to decompose the raw data into a lower dimensional subspace, as illustrated in Figure \ref{fig:f1} and Equation \ref{eq:tucker3}. After the decomposition we obtained matrices A, B and C, a small core tensor G and a tensor of residual errors.

\begin{figure}[ht]
 \begin{center}
	\includegraphics[width=0.8\textwidth]{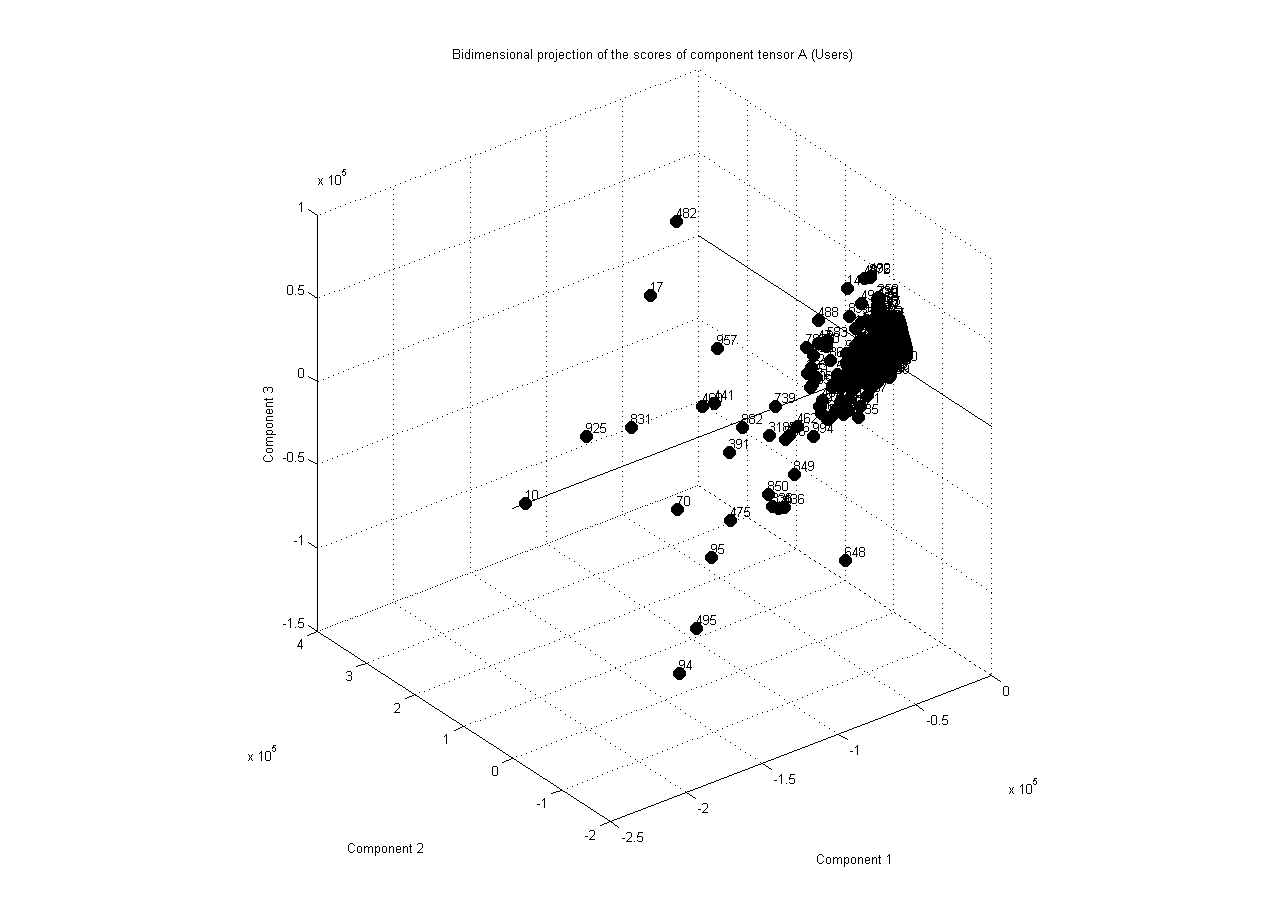}
 \end{center}
 \caption{Projection of Users on Matrix A for dataset X102} \label{fig:f3}
\end{figure}

In order to detect the abnormal users we simply projected the users on the component space yielded by matrix A. This projection is presented in Figure \ref{fig:f3}, for dataset X102. The three-dimensional subspace is given by the three obtained components by the model for the 1st mode (users). As mentioned earlier, this number of components is one of the parameters of the model, namely P=3, which corresponds to the first mode.

In order to evaluate the reliability of the model we used the same procedure and applied a Tucker3 model to dataset X909, which includes all users of X102. Our idea was to see how this model can identify abnormal users from both datasets. For this purpose, we computed the Euclidean distance between each user in the projection space (see Figure \ref{fig:f3}) and the corresponding center (0, 0, 0), for both datasets X102 and X909. Then we normalized the distances for each dataset and computed the Pearson correlation for the common users of these two datasets, according to their distance to the center of the subspace. We obtained a correlation of 68.44\%. Although, for X909 we just took 3 out of 40 main components to and model fit was different for both datasets (42\% for X102 and 51.01\% for X909),  abnormal or normal users in X102 approximately appeared as the same way in X909 with 68.44\% confidence. This denotes that Tucker3 is a robust model to detect the abnormal users.

\subsection{User Trajectories} \label{sec:usertrajectories}

\begin{figure}[ht]
 \begin{center}
	\includegraphics[width=0.5\textwidth]{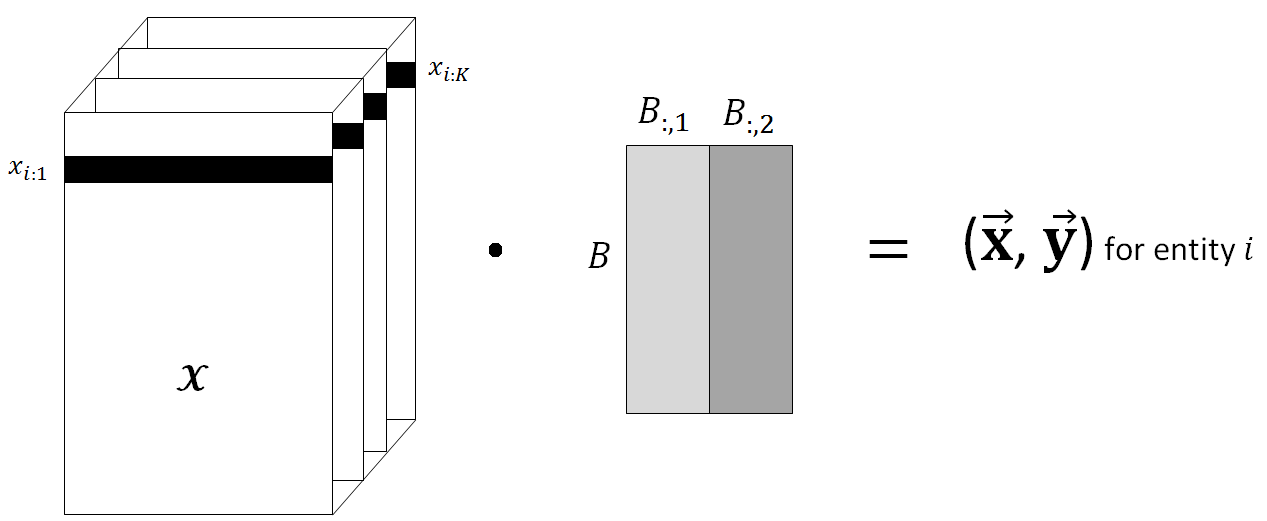}
 \end{center}
 \caption{Two sample users trajectories in X909, Left) 4th ranked abnormal user Right) 2nd ranked abnormal user} \label{fig:f4}
\end{figure}

Visualization methods like the one we presented in Figure \ref{fig:f3} are not able to show the evolving behavior of users over time.  We need another solution to enable us understanding the behavior of users over time. One solution is to project the users on a decomposed feature space (matrix B of Figure \ref{fig:f1}) for each time point. Since both of our selected parameters have Q equal to 2 it means that after projecting Users on feature space we must have a coordinate of (x,y) for each timepoint and for each user. The process of generating this coordinates is presented in Figure \ref{fig:f4}. $B(:,1)$ and $B(:,2)$ represent the two components that summarize the original entities of the features mode and X represent the three-order tensor (see Figure \ref{fig:f1}). The rows of the front matrix are the users, the columns correspond to the features and the third mode (z-axis) represents the hours. If we compute the dot product between each tensor's rows with the columns of the component matrix B, yielded by the Tucker3 model we obtain the coordinate(x,y) for a given time-point. If we repeat this procedure for all time points (e.g. hours), we are able to generate the coordinates of each user for the 720 hours. The user trajectories are obtained by sequentially connecting these coordinates. Formally we define user trajectories as: \\ 

\textbf{\textbf{Definition 1 (User Trajectory)}} : A sequence of time-stamped points, Trj= $trj: p_0 \rightarrow p_1 ... \rightarrow  p_i \rightarrow  ... p_k$, where $p_i (x,y,t) (i=0,1, ..., k)$, and $t$ is a time point.  \\

\begin{figure}[ht]
 \begin{center}
	\includegraphics[width=0.5\textwidth]{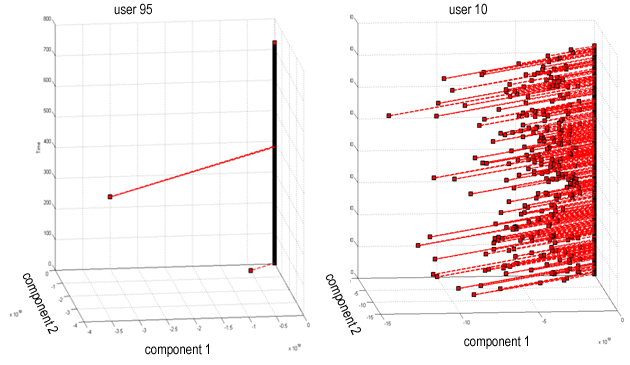}
 \end{center}
 \caption{Generation process of user trajectories} \label{fig:f5}
\end{figure}

Figure \ref{fig:f5} shows two abnormal users appearing in the top-10 users ranked based on abnormality. These abnormal users were ranked based on decreasing values of distance to the center, as explained in subsection 4.1, user 10 (right) is ranked 2nd and user 95 (left) is ranked 4th.  However, as it is clear from the figure, their behavior over time is completely different. User 95 just shows two abnormal behaviors that correspond to two time points, while user 10 shows this abnormal behavior almost in all time points. This means that user 10 is dealing with a stable problem while user 95 only has problems in specific points in time. This type of interpretation was not possible based only on the ranking list of abnormal users, obtained in subsection 4.1. Using user trajectories provides us richer insights into different kind of problems a user can experience.  For instance, what made user 95 be identified as abnormal could be something that suddenly happened in the network and then was quickly solved, while for user 10, some problems occurred but they were not solved until the end of the time period under analysis.

\subsection{Event Detection from user trajectories} \label{sec:eventdetection}

Even though user trajectories can be useful, when the number of users is too large, the individual analysis of each trajectory can become a cumbersome task. If we notice that some group of users trajectories behave similarly, this can be understood as something abnormal happens in their network level. Then some prevention or surveillance operations can be conducted more quickly. 

\begin{figure}[ht]
 \begin{center}
	\includegraphics[width=0.7\textwidth]{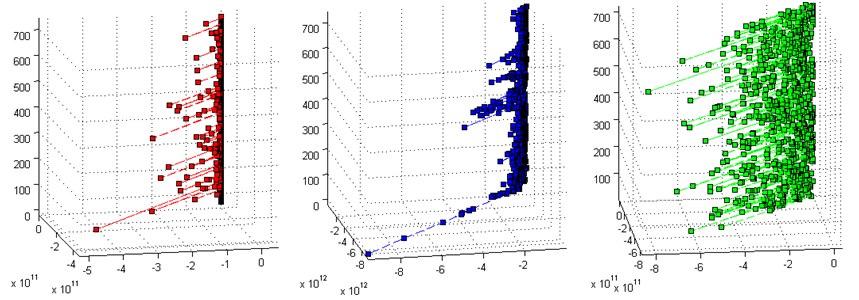}
 \end{center}
 \caption{Center Trajectory of clusters, Left: 1 user, Center : 866 users and Right: 22 users.} \label{fig:f6}
\end{figure}

To explore this goal, we employed Agglomerative Hierarchical Clustering toolbox from MATLAB to cluster user trajectories. We defined Euclidean distance between each point in trajectories as our distance function and Ward's criterion as the linkage criterion. We tested different values of cut-off from 0.6 to 1.2 to examine the clustering structure. The most suited clustering structure was obtained for a dendrogram distance of 1, which cuts the tree to level that, corresponds to three clusters. The average trajectory of these clusters is shown in Figure \ref{fig:f6}. Cluster red has1 user (0.1\%), cluster blue comprises 866 users (97.4\%) and cluster green includes 22 users (2.5\%). As it can be seen, no specific pattern can be recognized from the green and the red cluster. The users in these two clusters show an abnormal behavior almost in all time points. Such event can be due to a stable specific problem such as a problem in the user device. 

Regarding the blue cluster, it is possible to detect three events. First significant event occurs between hours 350 to 400. Second and third events also occur between 450 to 480 and 520 to 560, respectively.  However, the occurrence of the second and the third events should be assessed with hypothesis testing since they can be due to an accidental change.

\section{Conclusion and Future works} \label{sec:conclusion} 

In this paper, we present a study on using the Tucker3 decomposition to discover abnormal users in an IP/TV network. Our results indicate that Tucker3 is a robust method for detecting abnormal users in situations where interactions between the three dimensions are present.  From the tensor decomposition, we can define user trajectories. The trajectories allow us to observe the behavior of these users over time. We were able to identify two kinds of abnormal users: those who show frequent abnormal behavior over the whole time period and those who are associated to one or few severe abnormal behaviors over the time period. Without resorting to the analysis of user temporal trajectories it would have been harder to uncover such facts. Furthermore, from the clusters of the users trajectories, we have identified three events that occurred during three time points in the network. The result of this work can be used in a real network surveillance system to identify failures in the quickest possible time.

In this work, we did not consider the spatial relation of users.  Taking into account spatial relationships between network nodes could lead to a better clustering of users. Since some users might show similar behavior, with some delays, other distance measures for clustering should be tested. Currently we are employing another distance function using dynamic time warping, which assigns two users with same behavior but with a time shift in the same cluster.  The solution we presented for detection of events was based on clustering of trajectories. We are going to apply sliding window on trajectories to find time periods which have the most compact trajectories, which would lead to the discovery of events in a more accurate and reliable way.

\subsubsection*{Acknowledgments.} This work was supported by the Institute for Systems and Computer Engineering of Porto (INESC TEC) under projects TNET with reference BI/120033/PEst/LIAAD and project KDUS with reference  PTDC/EIA-EIA/098355/2008. The authors are grateful for the financial support.

\bibliographystyle{chicago}
\bibliography{ref}

\end{document}